\documentclass[journal]{IEEEtran}
\usepackage{cite}
\usepackage{times}
\usepackage{epsfig}
\usepackage{graphicx}
\usepackage{amsmath}
\usepackage{amssymb}
\usepackage{caption}
\usepackage{graphicx}
\usepackage{amssymb}
\usepackage{array}
\usepackage{booktabs}
\usepackage{multirow}
\usepackage{color}
\usepackage[ruled, linesnumbered]{algorithm2e}

\usepackage{comment}
\usepackage{amsmath,amssymb} 
\usepackage{color}
\usepackage{graphicx}
\usepackage{subcaption}
\usepackage{booktabs}
\usepackage{multirow}
\usepackage[ruled, linesnumbered]{algorithm2e}
\usepackage{algpseudocode}
\usepackage{bbding}
\usepackage{wrapfig}
\usepackage{picinpar}

\SetKwProg{Init}{Init}{}{}
\SetKw{And}{and}

\usepackage[colorlinks,
            linkcolor=red,
            anchorcolor=blue,
            citecolor=green
            ]{hyperref}

\usepackage[capitalize]{cleveref}
\crefname{section}{Sec.}{Secs.}
\Crefname{section}{Section}{Sections}
\Crefname{table}{Table}{Tables}
\crefname{table}{Tab.}{Tabs.}

\usepackage{xspace}
\newcommand{\name}[0]{TraSw\xspace}

\makeatletter
\DeclareRobustCommand\onedot{\futurelet\@let@token\@onedot}
\def\@onedot{\ifx\@let@token.\else.\null\fi\xspace}
\def\eg{\emph{e.g}\onedot} 
\def\ie{\emph{i.e}\onedot} 
 
\def\etc{\emph{etc}\onedot} 
 
\def\etal{\emph{et al}\onedot}

\begin{document}
%
\title{Tracklet-Switch Adversarial Attack against Pedestrian Multi-Object Tracking Trackers}
%
%
%

\author{
Delv~Lin$^*$,~Qi~Chen$^*$,~Chengyu~Zhou,~and~Kun~He$^\dag$
\IEEEcompsocitemizethanks{\IEEEcompsocthanksitem The authors are with Huazhong University of Science and Technology, Hubei, China.
E-mail: \{derrylin,~bloom24,~hust\_zcy,~brooklet60\}@hust.edu.cn.}
\thanks{The first two authors contribute equally. }
\thanks{Corresponding author: Kun He.}
}
\maketitle

\begin{abstract}
Multi-Object Tracking (MOT) has achieved aggressive progress and derived many excellent deep learning trackers. Meanwhile, most deep learning models are known to be vulnerable to adversarial examples that are crafted with small perturbations but could mislead the model prediction. 
In this work, we observe that the robustness on the MOT trackers is rarely studied, and it is challenging to attack the MOT system since its mature association algorithms are designed to be robust against errors during the tracking. To this end, we analyze the vulnerability of popular MOT trackers and propose a novel adversarial attack method called Tracklet-Switch (TraSw) against the complete tracking pipeline of MOT. The proposed TraSw can fool the advanced deep pedestrian trackers (\ie, FairMOT and ByteTrack), causing them fail to track the targets in the subsequent frames by perturbing very few frames. Experiments on the MOT-Challenge datasets (\ie, 2DMOT15, MOT17, and MOT20) show that TraSw can achieve an extraordinarily high success attack rate of over 95\% by attacking only four frames on average. To our knowledge, this is the first work on the adversarial attack against the pedestrian MOT trackers. Code is available at \url{https://github.com/JHL-HUST/TraSw}.
\end{abstract}


%
\IEEEpeerreviewmaketitle

\section{Introduction}
\label{sec:intro}
\begin{figure*}[t]
  \centering
   \includegraphics[width=0.9\linewidth]{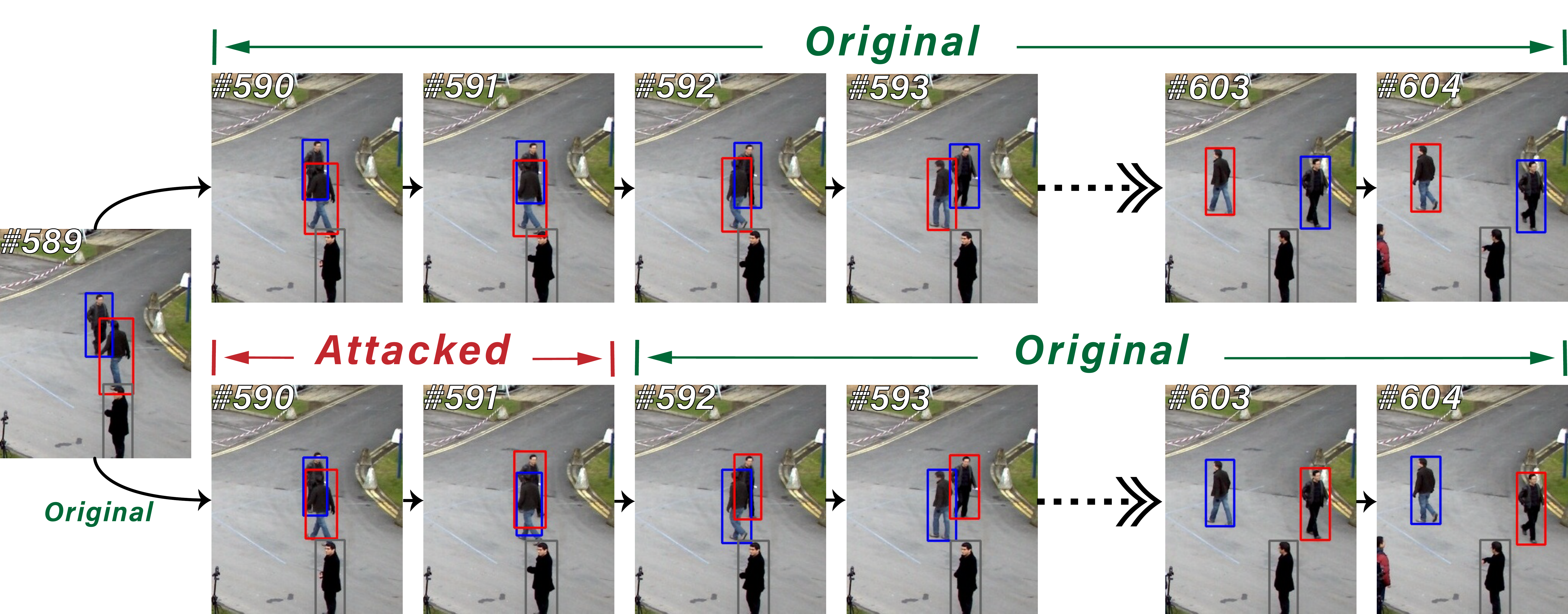}
   \caption{\footnotesize The illustration of the adversarial effect to tracking. \textit{First row}: the original video clips. \textit{Second row}: the adversarial video clips. The blue boxes represent the detected objects of the 19th trajectory, the red boxes represent those of the 24th trajectory, and the gray boxes represent the 16th trajectory not participated in the attack.
   Attacking two frames in the example video, \name switches the trajectories of the 19th and the 24th. After the attack, the exchanged state can hold until the end of the video.}
   \label{fig:f1}
\end{figure*}

\IEEEPARstart{O}{wing} to the rapid development of Deep Neural Networks (DNNs), Visual Object Tracking (VOT)~\cite{wojke2017simple,chen2018real} has been dramatically boosted in recent years with a wide range of applications, such as autonomous driving~\cite{li2013survey}, intelligent monitoring~\cite{xu2018real}, human-computer interaction~\cite{candamo2009understanding}, \etc. 
Researches on VOT fall into two categories, namely Single-Object Tracking (SOT) and Multi-Object Tracking (MOT). The goal of SOT is to track 
a single target given in the first frame of the video, while MOT aims to track all the targets of interest in an unsupervised setting and link the objects in different frames to form their trajectories.

On the other hand, deep learning models of various computer vision tasks are known to be vulnerable to adversarial examples~\cite{szegedy2013intriguing,goodfellow2014explaining,kurakin2018adversarial,eykholt2018robust,zhao2019seeing}, which are crafted with imperceptible perturbations but lead the models to wrong predictions. 
Studying the generation of adversarial examples can help promote the understanding of internal mechanism on DNNs, dig out their potential risks, and assist us to design robust deep learning systems. 
Therefore, adversarial attacks have been extensively studied in many computer vision tasks, such as image classification~\cite{nguyen2015deep,szegedy2014intriguing,moosavi2017universal,kurakin2018adversarial,eykholt2018robust,lin2020nesterov}, object detection~\cite{lu2017adversarial,bao2020sparse,10.1007/978-3-030-10925-7_4,hu2021naturalistic}, semantic segmentation~\cite{xie2017adversarial,hendrik2017universal,DBLP:conf/iclr/FischerKMB17}, \etc.
Current adversarial attack methods of VOT are mainly concentrated in Single-Object Tracking~\cite{chen2020one,yan2020cooling,jia2021iou}, and there is little work on attacking the Multi-Object Tracking system. To our knowledge, there exists only one work that does MOT adversarial attack on autonomous driving~\cite{jia2020fooling}. 

In this work, we focus the adversarial attacks on pedestrian tracking and investigate the robustness of these MOT trackers. The underlying reasons for this specification are twofold. First, the previous adversarial study, \textit{tracker hijacking}, is focused on vehicle tracking~\cite{jia2020fooling}, in which the attacked models are the detection-based trackers. But motion-and-appearance-based trackers~\cite{yu2016poi,li2019multiple,sommer2021appearance} are also popular and the performance of the detection-based trackers has been improved significantly over the years. In Table~\ref{tab:t1}, we observe that for the recent state-of-the-art detection-based tracker, ByteTrack~\cite{zhang2021bytetrack}, the attack success rate of \textit{tracker hijacking} is much lower than that of the method proposed in this paper, especially under the crowded tracking scenarios.
Second, most of the current MOT researches focus on pedestrian tracking. According to the statistics in~\cite{luo2020multiple}, at least 70\% of the current MOT research efforts are devoted to pedestrians, but there is no  corresponding adversarial study. 

A typical MOT tracker addresses the tracking in two steps~\cite{wang2020towards,centertrack,luo2020multiple}. First, the tracker locates all the objects in each frame. Then, according to the metric of similarity, each detected object is associated with a trajectory. It is important and challenging to consistently label the objects over time, especially in complex scenes with occlusion and interaction of objects. Great efforts have been devoted to improve the tracking robustness~\cite{bewley2016simple,yoon2016online,yu2016poi,wang2020towards,zhang2021fairmot}. 

Attacking these trackers poses new challenges due to their memory characteristics. Specifically, the continuous tracking process allows the tracker to save the moving states of the trajectories for a long period (\eg, 30 frames). As a result, if an object disappears for a few frames, it actually has little impact on the final trajectory. Moreover, the corresponding trajectory will be removed only if the tracking object disappears for sufficient frames (\eg, 30 frames).
Simply removing or tampering with the objects is quite inefficient. In~\cref{fig:4}, we find that even if the attacked frames are up to 20, the attack success rate of the object detection attacks that hide objects is hard to reach 90\%.

To address the above challenges, we propose a novel attack method called the \textit{Tracklet-Switch} (\name). In a nutshell, our method learns an effective perturbation generator to make the tracker confuse intersecting trajectories by attacking as few as one frame, as illustrated in \cref{fig:f1}, and the error state can transfer across frames until the end of the sequences. 
 In our method, we propose two novel losses, \textit{PushPull} and \textit{CenterLeaping}. 
 \textit{PushPull} works on the re-identification (re-ID) branch, while \textit{CenterLeaping} works on the detection branch. They can work together or separately to accommodate different kinds of MOT trackers (\eg, detection-based trackers, motion-and-appearance-based trackers).
 
To better illustrate the effectiveness and efficiency of \name, we choose three MOT-Challenge datasets, \ie, 2DMOT15, MOT17 and  MOT20~\cite{2015motchallenge,milan2016mot16,dendorfer2020mot20}, for evaluation, and compare \name with representative object detection adversarial attack methods~\cite{lu2017adversarial,yan2020cooling,guospark,chen2020one}, as well as the MOT adversarial attack method, \textit{tracker hijacking}~\cite{jia2020fooling}. Experiments show that our method achieves a significantly higher success rate with fewer attacked frames and smaller perturbations.

\section{Related Work}
\label{sec:rw}
\subsection{Multi-Object Tracking}

Multi-Object Tracking aims to locate and identify the targets of interest in the video and estimate their movements in the subsequent frames~\cite{bewley2016simple,luo2020multiple,wang2020towards,zhang2021fairmot,sun2022dancetrack,giancola2018soccernet,cai2022memot,li2022learning}, such as pedestrians on the street, vehicles on the road, or animals on the ground. The mainstream MOT trackers break the tracking into two steps: 1) the detection stage, where objects are located in the images, 2) the association stage, where the objects are linked to the trajectories.

The MOT trackers are expected to keep consistent track of each object over time; that is, each object should be assigned a unique track ID which stays constant throughout the whole sequence. To ensure as accurate tracking as possible, the tracker uses tracklets to maintain the moving states of the existing trajectories (e.g., position, velocity, and appearance). For a coming frame, the tracker compares the detected objects with tracklets to determine whether it belongs to an existing trajectory or is a new trajectory. The matching process between the tracklets and the detected objects is regarded as a bipartite matching problem based on the pair-wise similarity affinity matrix between all the tracklets and detected objects~\cite{jia2020fooling}. Most trackers use the motion information for evaluation. The commonly used motion similarity evaluation metric is Intersection-over-Union (IoU), which measures the spatial overlapping between two bounding boxes (bboxes). Some trackers also calculate the appearance affinity matrix to measure their appearance similarity. For those trackers, that only utilize motion flows to track objects, we call them detection-based trackers~\cite{bewley2016simple,wang2021track,zhang2021bytetrack}. And for those using both motion and deep appearance feature of the object for 
matching, we call them motion-and-appearance-based trackers~\cite{yoon2016online,yu2016poi,wang2020towards}.

After that, the matched tracklets will be updated (\eg, motion state, appearance state), and the unmatched objects will be initialized as new trajectories. The unmatched tracklets won't be removed immediately, instead they are moved to the lost pool. If objects of the lost tracklets reappear, the corresponding tracklets will be updated and put back into the tracking pool. Otherwise, if the missing tracklet reaches the maximum cache period (\eg, 30 frames), the unmatched tracklet will be completely removed. Even if the target reappears later, it will be regarded as an entirely new trajectory.

\subsection{Adversarial Attacks on Visual Object Tracking}
{\bf Adversarial Attacks on Single-Object Tracking.} 
The SOT tracker is informed of the tracking target in the first frame, and the goal is to predict the position and size of the tracking target in subsequent frames~\cite{li2018high,li2019siamrpn++,zhang2019deeper}.

Since the tracking template is given in the first frame, if the tracking template is wrong, the tracker cannot track the target. 
The early SOT attack method, one-shot attack~\cite{chen2020one}, aims to disturb the template by adding perturbations on the template patch in the initial frame. To get the perturbation, the attacker needs to iterate over every frame in the video to compute the perturbation. However, traversing all the frames is inefficient and unpractical.

Subsequently, Yan \etal~\cite{yan2020cooling} propose a \textit{cooling-shrinking} attack method to fool the SiamPRN-based trackers~\cite{li2019siamrpn++,li2018high} by training a perturbation generator to craft noise to interfere the search regions so as to make the target invisible to the trackers. 
More recently, Jia \etal~\cite{jia2021iou} present a decision-based black-box attack method, called IoU attack, that aims to gradually decrease the IoU score between the bbox of the clean image and the bbox of the adversarial sample, which leads the prediction deviating from its original trajectory.
These methods need to attack all the video frames. As the attacked target is determined in SOT, the number of attacks is relatively small, and valid attack samples can be obtained quickly. However, in MOT, the tracking quantity and tracking template vary along the frames. 
It is difficult to obtain effective adversarial examples by simply transplanting the SOT attack methods.

{\bf Adversarial Attacks on Multi-Object Tracking.} Compared with SOT, there is little work on attacking MOT. 
There is only one MOT attack method proposed on autonomous driving, called \textit{tracker hijacking}~\cite{jia2020fooling}. 
By fabricating the bboxes toward the expected attacker-specified direction and erasing the original bboxes, the attacked object deviates from its original trajectory. As a result, the successful adversaries can move an object out of (into) the view of an autonomous vehicle. \textit{Tracker hijacking} attacks those detection-based trackers~\cite{luo2020multiple,jia2020fooling}, that crafts adversarial examples to fool the object detection module.

For reference, we provide a comparison of \textit{tracker\ hijacking} and our proposed \name on the stat-of-the-art pedestrian trackers.

\section{Preliminaries}
\label{sec:pre}
\begin{figure*}[ht]
  \centering
  \begin{subfigure}[t]{0.4\linewidth}
  \centering
    \includegraphics[width=0.9\linewidth]{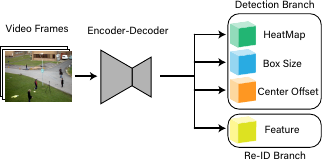}
    \caption{FairMOT network}
    \label{fig:f3a}
  \end{subfigure}
  \begin{subfigure}[t]{0.47\linewidth}
  \centering
    \includegraphics[width=0.9\linewidth]{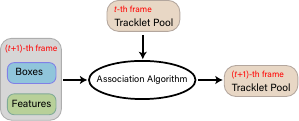}
    \caption{Association}
    \label{fig:f3b}
  \end{subfigure}
   \caption{Overview of FairMOT, a typical MOT tracker as our main target model. 
   }
   \label{fig:f3}
\end{figure*}

In this work, we choose FairMOT~\cite{zhang2021fairmot} and ByteTrack~\cite{zhang2021bytetrack} as our main target models for attack. 
As FairMOT and ByteTrack share similar structures expect for the re-ID branch in FairMOT, we primarily focus on introducing FairMOT to illustrate our proposed method.
In this section, we first introduce the overview of FairMOT, then we provide the problem definition of adversarial attack on MOT pedestrian trackers.

\subsection{Overview of FairMOT}
\label{fm}

FairMOT stands out among many trackers by achieving a good trade-off between accuracy and efficiency. As shown in \cref{fig:f3}, FairMOT consists of two homogeneous branches for object detection and feature extraction, and follows the standard online association algorithm.

{\bf Detection Branch.}
The anchor-free detection branch of FairMOT is built on CenterNet~\cite{duan2019centernet}, consisting of the heatmap head, box-size head and center-offset head. Denote the ground-truth (GT) bbox of the $i$-th object in the $t$-th frame as $box^i_t=(x_1^{t,i},y_1^{t,i},x_2^{t,i},y_2^{t,i})$. The $i$-th object's center $(c_x^{t,i},c_y^{t,i})$ is computed by  $c_x^{t,i}=\frac{x_1^{t,i}+x_2^{t,i}}{2}$ and $c_y^{t,i}=\frac{y_1^{t,i}+y_2^{t,i}}{2}$.
The response location of the center on the heatmap can be obtained by dividing the stride (which is 4 in FairMOT)  $(\lfloor\frac{c_x^{t,i}}{4}\rfloor,\lfloor\frac{c_y^{t,i}}{4}\rfloor)$. The heatmap value indicates the probability of  presence of an object centering at this point. The GT box size and center offset are computed by $(x_2^{t,i}-x_1^{t,i},y_2^{t,i}-y_1^{t,i})$ and $(\frac{c_x^{t,i}}{4}-\lfloor\frac{c_x^{t,i}}{4}\rfloor,\frac{c_y^{t,i}}{4}-\lfloor\frac{c_y^{t,i}}{4}\rfloor)$, respectively.

{\bf Re-ID Branch.}
The re-ID branch generates the re-ID features of the objects. Denote the feature map as $feat_t\in\mathbb{R}^{512\times H\times W}$. The re-ID feature $feat_t^i\in\mathbb{R}^{512}$ represents the feature vector of the $i$-th object, whose $L_2$ norm equals $1$.

{\bf Association.}
FairMOT follows the association strategy in~\cite{wang2020towards}.
The tracker uses tracklet to describe the trajectory's appearance state $a_t^i$ and motion state $m_t^i = (x^{t,i},y^{t,i},\gamma^{t,i},h^{t,i}, \dot{x}^{t,i},\dot{y}^{t,i},\dot{\gamma}^{t,i},\dot{h}^{t,i})$ in the $t$-th frame, where $(x^{t,i},y^{t,i})$ indicates the predicted center, $h^{t,i}$ the height and $\gamma^{t,i}$ the aspect ratio, followed by their velocity information. The initial appearance state $a_{0}^i$ is initialized with the first observation's
appearance embedding $feat_{0}^i$ of object $i$, and $a_{t}^i$ is updated by:
\begin{equation}
\label{eq:smooth feat}
\begin{aligned}
a_t^i = \alpha\cdot a_{t-1}^i + (1-\alpha)\cdot feat_t^i,
\end{aligned}
\end{equation}
where $feat_t^i$ is the appearance embedding of the matched object in the $t$-th frame. 
The bbox information in the motion state $m_t^i$ is updated by 
the predicted center $(x^{t,i},y^{t,i})$, height $h^{t,i}$ and aspect ratio $\gamma^{t,i}$ in the $t$-th frame, and velocity information $(\dot{x}^{t,i},\dot{y}^{t,i},\dot{\gamma}^{t,i},\dot{h}^{t,i})$ is updated by the Kalman filter. 

For a coming frame, we compute the pair-wise similarity matrix between the observed objects in the $t$-th frame and the tracklets maintained by the tracklet pool $TrP_{t-1}$. Then the association problem is solved by the Hungarian algorithm using the final cost matrix:
\begin{equation}
\label{eq:distance}
\begin{aligned} 
d_{t}=\lambda\cdot d_{box}(K(m_{t-1}),box_t)+(1-\lambda)\cdot d_{feat}(a_{t-1},feat_t),
\end{aligned}
\end{equation}
where $box_t$ and $feat_t$ denote the detected bboxes and features in the $t$-th frame. $K(\cdot)$ represents the Kalman filter that uses motion state $m_{t-1}$ to predict the expected positions of the trajectories in the $t$-th frame, and $d_{box}(\cdot)$ stands for a certain measurement of the spatial distance, which is the Mahalanobis distance in FairMOT, and $d_{feat}(\cdot)$ represents the cosine similarity.

\subsection{Problem Definition}
\label{sec:pd}

Let $V=\{I_1,\dots,I_t,\dots,I_N\}$ denote the sequence frames of a video. 
Consider a scenario where the tracker detects two trajectories intersected at frame $t$.
Denote the two trajectories as $T_i=\{O_{s_i}^i,\dots,O_t^i,\dots,O_{e_i}^i\}$ and $T_j=\{O_{s_j}^j,\dots,O_t^j,\dots,O_{e_j}^j\}$.
Their bboxes and features are  $B_k=\{box_{s_k}^k,\dots,box_t^k,\dots,box_{e_k}^k\}$ and $F_k=\{feat_{s_k}^k,\dots,feat_t^k,\dots,feat_{e_k}^k\}$ where $k\in\{i,j\}$, $box_t^k\in\mathbb{R}^4$ and $feat_t^k\in\mathbb{R}^{512}$. 

Then we define the adversarial video as 
$\widehat V=\{I_1,\cdots,I_{t-1},\widehat I_t,\cdots,\widehat I_{t+n-1},I_{t+n},\cdots,I_N\}$,
where $I$, $\widehat I$ indicate the original frame and adversarial frame, respectively. 
For the attack trajectory $T_i$, we call $T_j$, that overlaps with $T_i$ in the $t$-th frame, the screener trajectory. The adversarial video $\widehat V$ misleads the tracker to estimate trajectory $i$ as $\widehat T_i=\{O_{s_i}^i,\dots,O_{t-1}^i,O_t^j,\dots,O_{t+n-1}^j,O_{t+n}^j,\dots,O_{e_j}^j\}$. 
The goal is to attack the frame sequences to make the tracking of trajectory $T_i$ change to that of $T_j$ since the $t$-th frame.

\section{The \name Adversarial Attack}
\label{sec:method}

In this section, we propose a novel method called the \textit{Tracklet-Switch} (\name) adversarial attack against the pedestrian trackers, which aims to switch the tracklets of two intersecting trajectories in the video frames.
 In \cref{sec:pp} and \cref{sec:CL}, we propose the \textit{PullPush} loss and the \textit{CenterLeaping} technique. Then, we present the overall pipeline of \name in \cref{sec:gaa}. 


\subsection{Feature Attack with Push-Pull Loss}
\label{sec:pp}

\begin{figure*}[t]
  \centering
  \begin{subfigure}[b]{0.3\linewidth}
    \includegraphics[width=1\linewidth]{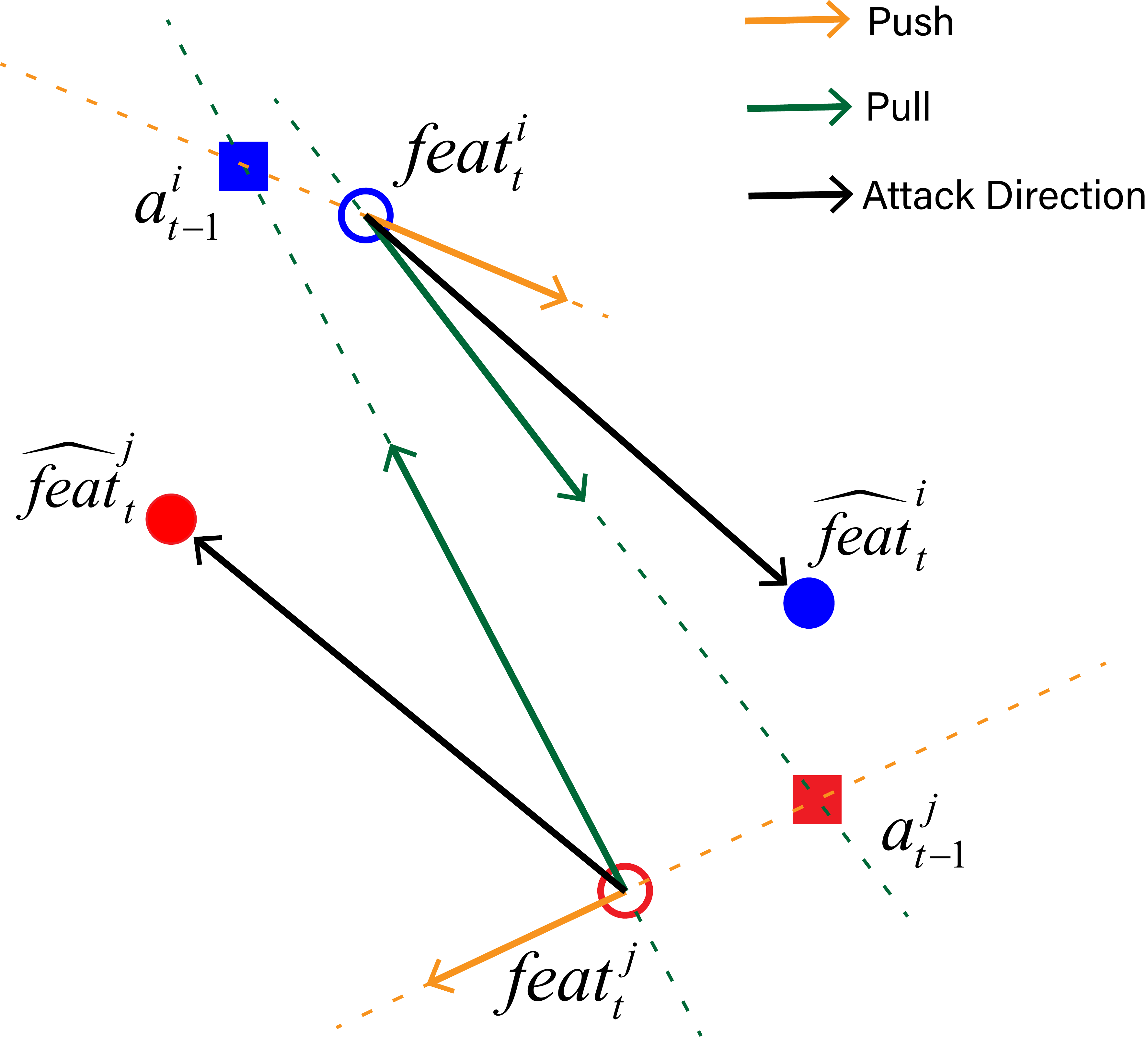}
    \caption{\footnotesize Push-pull loss}
    \label{fig:f4a}
  \end{subfigure}
  \hfill
  \begin{subfigure}[b]{0.26\linewidth}
    \includegraphics[width=1\linewidth]{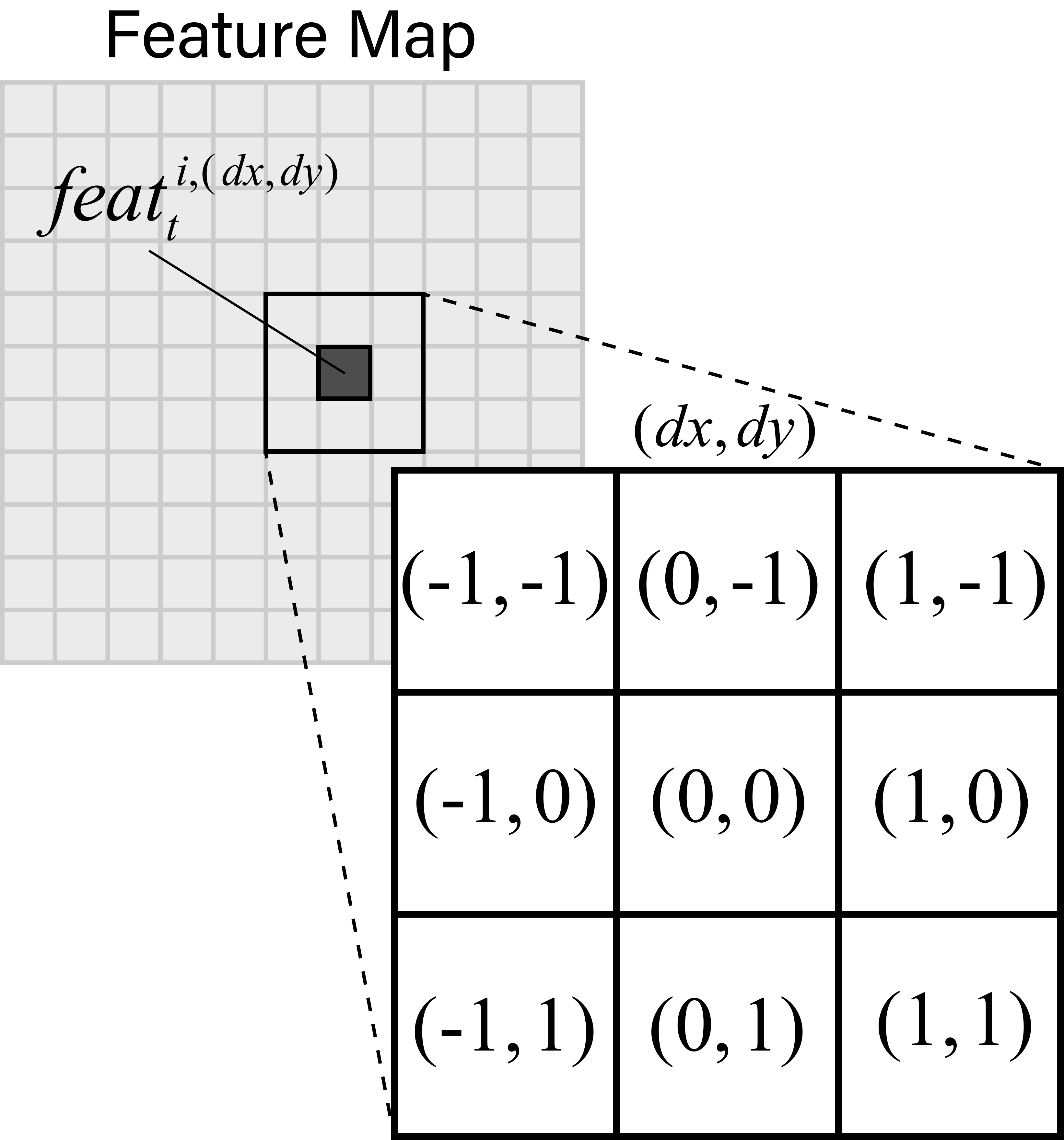}
    \caption{\footnotesize Nine-block box}
    \label{fig:f4b}
  \end{subfigure}
  \hfill
  \begin{subfigure}[b]{0.32\linewidth}
    \includegraphics[width=1\linewidth]{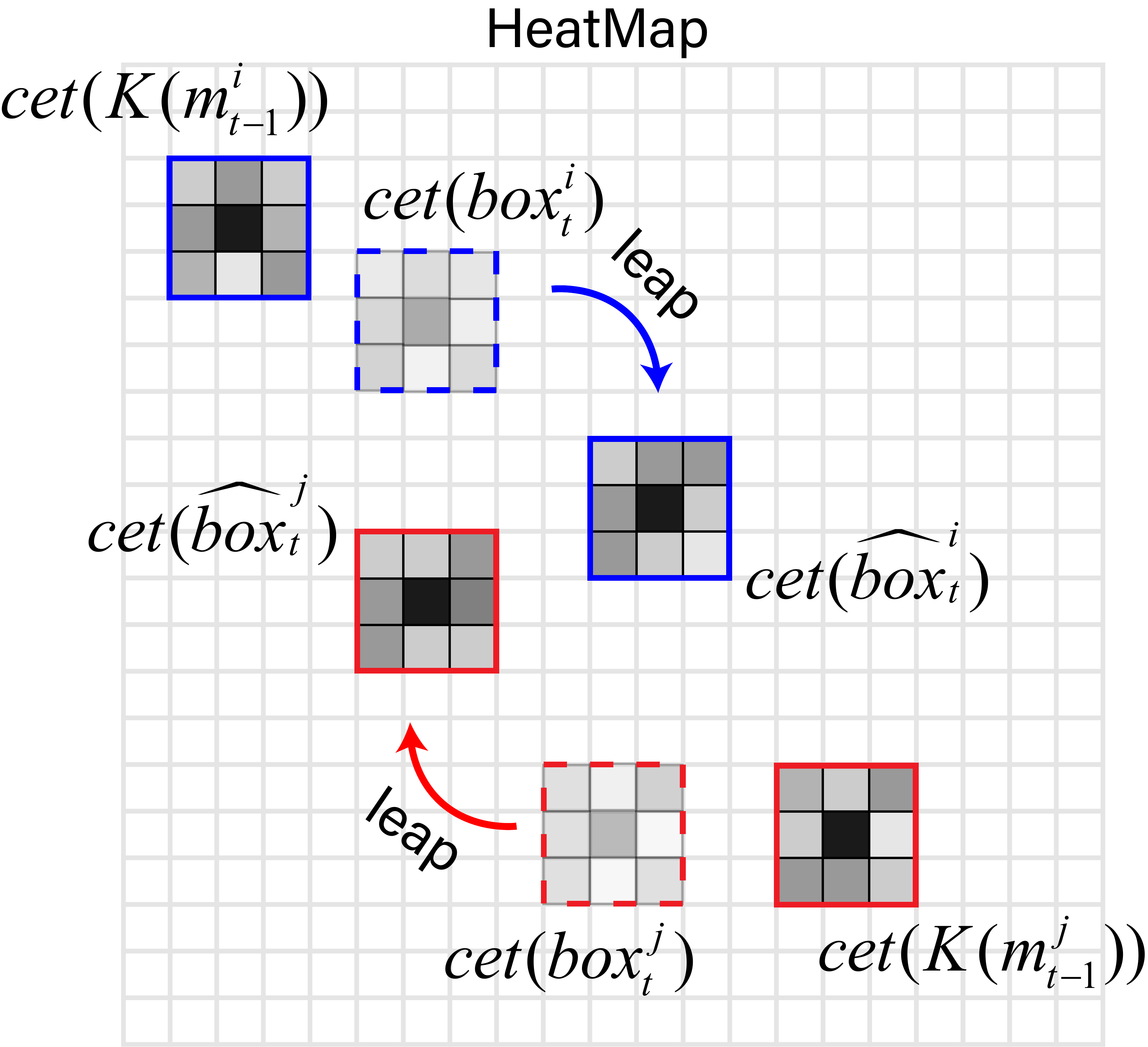}
    \caption{\footnotesize Center leaping}
    \label{fig:f4c}
  \end{subfigure}
  \caption{\footnotesize The key components of \name.
  (a) The illustration of $PushPull$ loss. The blue and red points represent different IDs to be attacked. The goal of the $PushPull$ loss is to push the feature $feat_t^i$ of object $i$ in frame $t$ away from that of tracklet $i$, $a_{t-1}^i$, and to pull $feat_t^i$ to be close to another tracklet $a_{t-1}^j$; and vise versa to $feat_t^j$. 
  (b) The nine-block box ($\mathcal{B}$). The surrounding features are also used to calculate the loss. 
  (c) The center leaping for the detection branch. The dotted boxes indicate the original heat points, which need to be cooled down; and the solid boxes in the $t$-th frame indicate the adversarial heat points, which need to be heated up. In this way, the $cet(box_t^i)$ and $cet(box_t^j)$ leaps to $cet(\widehat {box}_t^i)$ and $cet(\widehat {box}_t^j)$ along the direction of $cet(K(m_{t-1}^j))$ and $cet(K(m_{t-1}^i))$.}
  \label{fig:short}
\end{figure*}

The tracker distinguishes the objects through a combination of motion and appearance similarity. When objects are close to each other, the tracker relies more on the re-ID features to distinguish the objects. Therefore, we can reform the re-ID features of the objects to make them similar to another tracklet. Inspired by the triplet loss~\cite{schroff2015facenet}, we design the \textit{PushPull} loss  
to push away the attack (screener) feature and pull the screener (attack) feature, as follows: 
\begin{equation}
\begin{aligned}
\mathcal{L}_{pullpush}(a_{t-1}^i,a_{t-1}^j,feat_t^i,feat_t^j)=\\
\sum_{k\in\{i,j\}}
d_{feat}(a_{t-1}^k,feat_t^{\widetilde k}) - d_{feat}(a_{t-1}^k,feat_t^k),
\end{aligned}
\end{equation}
where $d_{feat}(\cdot)$ denotes the cosine similarity, $a_{t-1}^i$ and $a_{t-1}^j$ represent the appearance feature of the attack and screener tracklets, $feat_t^i$ and $feat_t^j$ represent the feature of the attack and screener objects, and $k$ represents the attack (screener) ID while $\widetilde k$ represents the screener (attack) ID, whose object overlaps most with object $k$ (\ie, $\widetilde i=j$ and $\widetilde j=i$).
The loss will make $feat_t^k$ dissimilar to tracklet $k$ and make $feat_t^{\widetilde k}$ similar to tracklet $k$ (see \cref{fig:f4a}).

Specifically, in FairMOT, the object's feature is extracted from the feature map $feat_t\in\mathbb{R}^{512\times H\times W}$ according to the predicted object center $(x,y)$. Considering the surrounding locations of the center that may be activated, we calculate the appearance cost within a nine-block box location (as illustrated in \cref{fig:f4b}) for a more stable attack. So the final \textit{PushPull} loss for FairMOT is as follows:
\begin{equation}
\small
\begin{aligned}
\mathcal{L}_{pp}=\sum_{(dx,dy)\in \mathcal{B}}
\mathcal{L}_{pullpush}(a_{t-1}^{i},a_{t-1}^{j},feat_t^{i,(dx,dy)},feat_t^{j,(dx,dy)}),
\end{aligned}
\end{equation}
where $\mathcal{B}$ indicates a set of offsets in the nine-block box location, $feat_t^{i,(dx,dy)}$ and $feat_t^{j,(dx,dy)}$ represent the feature extracted around the center of the attack and screener object, respectively.

\subsection{Detection Attack with Center Leaping}
\label{sec:CL}
Attacking the features of intersection trajectories can generally fool the tracker. But it may be insufficient when the spatial distance is too large to switch, and some trackers only use the detection result for the association.

In order to adjust the distance between objects and tracklets, we propose an efficient method, called the \textit{CenterLeaping}. The optimization objective can be summarized as reducing the distance between the tracklet and the target object. As the bboxes are computed with discrete locations of heat points in the heatmap, we cannot directly optimize the loss of bboxes to make them close to each other. So the goal is achieved by reducing the distance between the centers of bboxes, as well as the differences of their sizes and offsets. Hence, the detection optimization function for the attack trajectory $k$ can be expressed as follows: 
\begin{equation}
\label{eq:eq7}
\begin{aligned}
\min\sum_{k\in\{i,j\}}&d_{box}(K(m_{t-1}^{\widetilde k}),{box}_t^{k})\\
=\min\sum_{k\in\{i,j\}}\Big(&d(cet(K(m_{t-1}^{\widetilde k})),cet({box}_t^{k}))+\\
&d(size(K(m_{t-1}^{\widetilde k})),size({box}_t^{k}))+\\
&d(off(K(m_{t-1}^{\widetilde k})),off({box}_t^{k}))\Big),
\end{aligned}
\end{equation}
where $m_{t-1}^{k}$ represents the motion state of the attack or screener tracklet, $box_t^k$ represents the bbox of the attack or screener object, $d(\cdot)$ denotes the $L_1$ distance, $cet(\cdot)$, $size(\cdot)$ and $off(\cdot)$ compute the bbox's center, size and offset, respectively.

To make the object to be close to the center of the target tracklet, based on the focal loss of FairMOT, we design the \textit{CenterLeaping} loss to let the center of the attack (screener) bbox move close to the screener (attack) trajectory (see \cref{fig:f4c}):
\begin{equation}
\label{eq:cl}
\begin{aligned}
\mathcal{L}_{cl}=\sum_{k\in \{i,j\}}
\Big(&\sum_{(x,y)\in \mathcal{B}_{c_{\nearrow{\widetilde k}}}}-(1-M_{x,y})^\gamma\log(M_{x,y})+\\
&\sum_{(x,y)\in \mathcal{B}_{c_k}}-(M_{x,y})^\gamma\log(1-M_{x,y})\Big),
\end{aligned}
\end{equation}
where $M_{x,y}$ represents the value of heatmap at location $(x,y)$, $c_k$ represents the center of the attack or screener object, and $c_{\nearrow{\widetilde k}}$ represents the point in the direction from $c_k$ to $cet(K(m_{t-1}^{\widetilde k}))$. 
During the optimization iteration, point $c_{\nearrow{\widetilde k}}$ will leap to the next grid along the direction. As a result, the heatmap values around the original object centers get cooled down, while the points close to $cet(K(m_{t-1}^{\widetilde k}))$ are warmed up.

We also restrain the sizes and offsets of the objects by a regression loss:
\begin{equation}
\begin{aligned}
\mathcal{L}_{reg}=&\mathcal{L}_{size}+\mathcal{L}_{offset}\\
=&\sum_{k\in \{i,j\}}\mathcal{L}_1^{smooth}(size(K(m_{t-1}^{\widetilde k})), size(box_t^k))+\\
&\sum_{k\in \{i,j\}}\mathcal{L}_1^{smooth}(off(K(m_{t-1}^{\widetilde k})),off(box_t^k)),
\end{aligned}
\end{equation}
where $\mathcal{L}_1^{smooth}$ denotes the smooth $L_1$ loss:
\begin{equation}
\mathcal{L}_1^{smooth}(a, b)=
\begin{cases}
0.5\cdot(a-b)^2 & \text{if}\ |a-b|<1,\\
|a-b|-0.5       & \text{else}.
\end{cases}
\end{equation}

\subsection{Crafting the Adversarial Video}
\label{sec:gaa}

\begin{algorithm*}[h]
\caption{The \name Attack}
\label{alg:a1}
\KwIn{Video image sequence $V=\{I_1,\dots,I_N\}$;
MOT $Tracker(\cdot)$; attack ID $ID_{att}$; attack IoU threshold $Thr_{IoU}$; start attack frame $Thr_{frame}$; maximum iteration $Thr_{iter}$}
\KwOut{Sequence of adversarial video $\widehat V$; original tracklet pool $TrP_N$; adversarial tracklet pool  $\widehat{TrP}_N$}
\Init{
$\widehat V \gets \{\}$;
$TrP_0 \gets None$;
$\widehat{TrP}_0 \gets None$
}{}
\For{t=1 to N}
{$TrP_t\gets Tracker(I_t,TrP_{t-1})$\;
$\widehat{TrP}_t\gets Tracker(I_t,\widehat{TrP}_{t-1})$\;
$\widehat I_t\gets I_t$;\Comment{initialize the outputs}\\
\Comment{check if the attack tracklet has been existed for at least $Thr_{frame}$ frames and the object is tracked as $ID_{att}$}\\
\If{$Exist(ID_{att},TrP_t) > Thr_{frame}$ \And
$CheckFit(TrP_t|ID_{att},\widehat{TrP}_t|ID_{att})$}
{\Comment{find the screener object that overlaps most with the attack object~~~~~~~}\\
$ID_{scr}\gets FindMaxIoU(TrP_t,ID_{att})$\; 
\If{$IoU(TrP_t|ID_{att},TrP_t|ID_{scr})>Thr_{IoU}$}
{\Comment{generate adversarial noise iteratively with \cref{eq:AE generate}}\\
$noise\gets NoiseGenerator(ID_{att},ID_{scr},I_t,TrP_t,\widehat {TrP}_{t-1},Tracker(\cdot),Thr_{iter})$\;
$\widehat I_t\gets Cilp_{[0,1]}(I_t+noise)$;\Comment{clip the adversarial image to $[0,1]$}\\
$\widehat {TrP}_t\gets Tracker(\widehat I_t, \widehat {TrP}_{t-1})$;\Comment{update the adversarial tracklet}\\
$Thr_{IoU}\gets 0$;\Comment{set $Thr_{IoU}$ to $0$ if start to attack}\\
}}
$\widehat V\stackrel{+}{\gets} \widehat I_t$; \Comment{update the adversarial video}\\
}
\end{algorithm*}

Summing up the above, we get the total loss for optimization: 
\begin{equation}
\label{eq:final loss}
\min_{\widehat V} Loss=\min_{\widehat V}~\mathcal{L}_{pp}+\mathcal{L}_{cl}+\mathcal{L}_{reg}.
\end{equation}
Then we can calculate the gradient of the total loss with an $L_2$ regularization:
\begin{equation}
\label{eq:AE generate}
\widehat I_0=I,\ \widehat I_{i+1}=Clip_{[0,1]}\left(\widehat I_{i}-\frac {\nabla_{\hat I} Loss(\widehat I_i;\theta)}{\parallel\nabla_{\hat I} Loss(\widehat I_i;\theta)\parallel_2}\right),
\end{equation}
where $\hat I_i$ denotes the adversarial image at the $i$-th iteration.

The algorithm overview of crafting the adversarial videos is shown in \cref{alg:a1}. 
First, we specify an attack trajectory $ID_{att}$ in the original tracking video before the attack. For each coming frame, we initialize the tracklet pools, $TrP_t$ and $\widehat{TrP}_t$, with the original frame, and initialize the adversarial frame $\widehat{I}_t$ as $I_t$. 
Second, we conduct a double check to determine whether to attack the current frame: 
1) check whether the object of trajectory $ID_{att}$ has appeared for more than $Thr_{frame}$ frames (10 in default) with $Exist(\cdot)$. This is because the overall attack will make no sense if the attack starts from the first appearance of the attack target;
2) check whether the tracking of the attack object is the same as the original video with $CheckFit(\cdot)$.
If both conditions are satisfied, we then try to find an object that overlaps most with object $ID_{att}$ as the screener object $ID_{scr}$ by $FindMaxIoU(\cdot)$. 
Then we check whether the IoU between objects $ID_{att}$ and $ID_{scr}$ is greater than $Thr_{IoU}$. 
If true, the current frame will be attacked, and otherwise not. 

For the attacking, we use the $NoiseGenerator(\cdot)$ to generate adversarial noise by optimizing~\cref{eq:final loss} iteratively. 
In particular, for the \textit{CenterLeaping} loss, when the number of iterations reaches 10, 20, 30, 35, 40, 45, 50 or 55, the heat point $c_{\nearrow{\widehat k}}$ will advance one grid towards $c_k$ from its current position. $c_{\nearrow{\widehat k}}$ won't stop advancing until it overlaps with $c_k$.
During each iteration of optimization and attack, a specific noise will be generated and added to the video frame by \cref{eq:AE generate} until the tracker makes mistakes in the current iteration or the number of iterations reach $Thr_{iter}$ (60 in default).

We add the noise to the current frame no matter whether the attack succeeds, and the experiments in~\cref{as} show that such operation contributes to an easier attack for the following frames. The tracklet pool $\widehat{TrP}_t$ is then re-updated by the adversarial frame $\widehat{I}_t$, and the threshold $Thr_{IoU}$ is set to zero. 
In the end, the adversarial frame $\widehat{I}_t$ is added to the adversarial video $\widehat{V}$.

\section{Experiments}
\label{sec:exp}
In this section, we first introduce the experimental setup, then we present the results of \name and the baselines to demonstrate the effectiveness and efficiency of \name. Next, we analyze the importance of components in \name through the ablation study. In the end, we discuss the influence of $Thr_{IoU}$ and present the distribution area of adversarial noise.

\subsection{Experimental Setup}

{\bf Target Models and Datasets.} We choose two representative 
trackers as the target models: FairMOT~\cite{zhang2021fairmot} and ByteTrack~\cite{zhang2021bytetrack}. FairMOT is also used for ablation studies. In particular, we only use \textit{CenterLeaping} to attack ByteTrack, as there is no re-ID branch in ByteTrack. We validate \name on the test sets of three benchmarks: 2DMOT15~\cite{2015motchallenge}, MOT17~\cite{milan2016mot16} and MOT20~\cite{dendorfer2020mot20}. 

{\bf Baselines.} \name is compared with three classic baselines:
1) random noise perturbation (denoted as RanAt) whose $L_2$ distance per frame is limited to $[2, 8]$ randomly; 
2) detection attack (denoted as DetAt) which aims to make the attack object invisible to the object detection module~\cite{lu2017adversarial,yan2020cooling,guospark,chen2020one} (commonly used in object detection and SOT attacks); 
3) \textit{tracker\ hijacking} attack (denoted as Hijack)~\cite{jia2020fooling}.

{\bf Evaluation Metric.} Similar to~\cite{jia2020fooling}, 
an attack is regarded as successful 
when the detected objects of the attack trajectory are no longer associated with the original tracklet after the attack. 
As described in~\cref{sec:gaa}, our method attacks when an object overlaps with the attack object. 
So the attack success rate depends on two factors. 
Firstly, we need to obtain the number of trajectories satisfying the attack conditions: 
1) the trajectory's object should have appeared for at least $Thr_{frame}$ (10 as the default value) frames;
2) there exists another object overlapping with the attack object, and the IoU should be greater than $Thr_{IoU}$ (0.2 as the default value).
Secondly, we need to obtain the number of successfully attacked trajectories for which the detected bboxes are no longer associated with the original tracklets after the attack. For comparison purposes, the attack conditions of the baselines are the same as \name. The effectiveness and efficiency of our method are demonstrated through the attack success rate ($Succ.\uparrow$), attacked frames ($\#Fm.\downarrow$), and $L_2$ distance ($L_2\downarrow$) per track ID of the successful attacks. To make the attack more efficient, the attacked frames are limited to at most 20.

\begin{figure*}[h]
  \centering
    \includegraphics[width=0.8\linewidth]{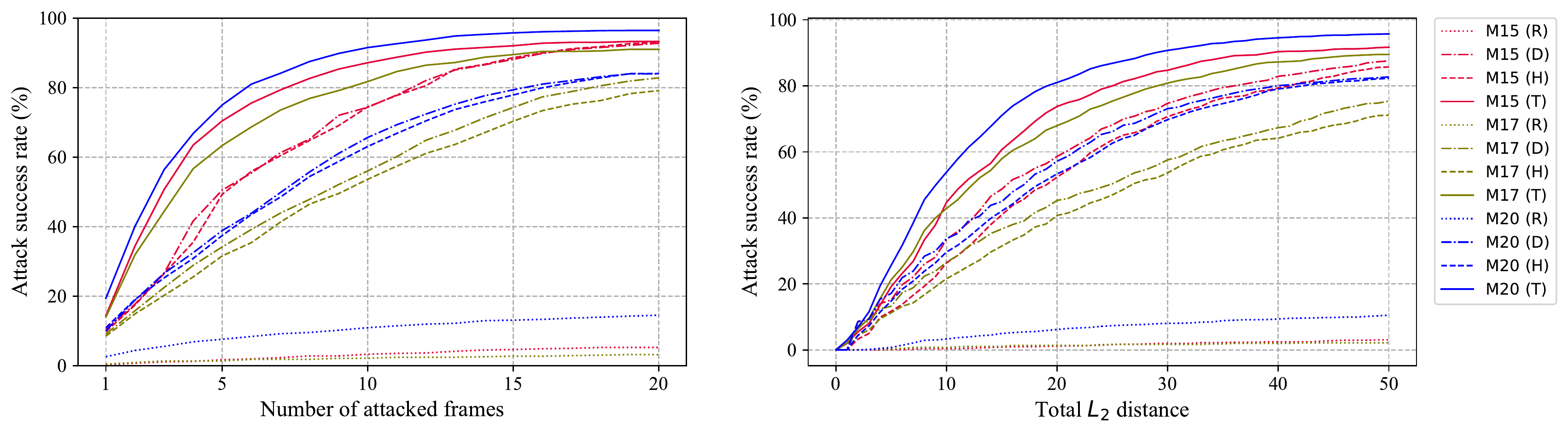}
  \caption{
  The evaluation results of trackers attacked by RanAt (R), DetAt (D), Hijack (H) and \name (T) on 2DMOT15 (M15), MOT17 (M17) and MOT20 (M20), respectively.
  }
  \label{fig:4}
\end{figure*}

\subsection{Adversarial Attack Results}

We report the results of \name and the baselines on the test datasets of 2DMOT15, MOT17, and MOT20. The results are as shown in Table~\ref{tab:t1}.

\begin{table*}[!ht]
\centering
\caption{Comparison of the attacks on the MOT-Challenge test datasets.}
\setlength{\tabcolsep}{2.5mm}{
\resizebox{0.8\linewidth}{!}{
\begin{tabular}[t]{c|c|c|ccc|c}
\toprule
Dataset & Tracker & Method & $Succ.\uparrow (\%)$ & $\#Fm.\downarrow$ & $L_2\downarrow$ & $IDs_{att}$ \\
\midrule
\multirow{8}* {2DMOT15}
&\multirow{4}* {FairMOT}
& RanAt & ~~5.25 & 8.74 & 43.85 & \multirow{4}* {820}\\
&& DetAt & 92.80 & 6.43 & 19.17 & \\
&& Hijack & 92.92 & 6.57 & 21.74 & \\
&& \name & \textbf{93.28} & \textbf{4.25} & \textbf{13.99} & \\
\cline{2-7}
&\multirow{4}* {ByteTrack} 
& RanAt & ~~4.80 & 6.83 & 34.14 & \multirow{4}* {730}\\
&& DetAt & 85.09 & 5.98 & 36.65 & \\
&& Hijack & 81.89 & 6.85 & 38.57 & \\
&& \name & \textbf{89.88} & \textbf{4.09} & \textbf{26.49} &\\
\midrule
\multirow{8}* {MOT17}
&\multirow{4}* {FairMOT}
& RanAt & ~~3.19 & 7.57 & 37.50 & \multirow{4}* {658}\\
&& DetAt & 82.83 & 7.84 & 23.07 & \\
&& Hijack & 79.18 & 8.00 & 24.39 & \\
&& \name & \textbf{91.03} & \textbf{4.74} & \textbf{14.40} & \\
\cline{2-7}
&\multirow{4}* {ByteTrack} 
& RanAt & ~~4.26 & 7.73 & 40.14 & \multirow{4}* {705}\\
&& DetAt & 70.35 & 6.04 & 27.81 & \\
&& Hijack & 67.53 & 6.95 & 32.31 & \\
&& \name & \textbf{91.06} & \textbf{4.17} & \textbf{24.02} & \\
\midrule
\multirow{8}* {MOT20}
&\multirow{4}* {FairMOT}
& RanAt & 14.53 & 6.77 & 34.17 & \multirow{4}* {1892}\\
&& DetAt & 83.99 & 6.81 & 16.31 & \\
&& Hijack & 84.04 & 7.09 & 17.88 & \\
&& \name & \textbf{96.46} & \textbf{3.94} & \textbf{11.67} & \\
\cline{2-7}
&\multirow{4}* {ByteTrack} 
& RanAt & ~~4.27 & 7.46 & 37.07 & \multirow{4}* {1899}\\
&& DetAt & 61.93 & 7.32 & 33.63 & \\
&& Hijack & 58.66 & 8.18 & 30.25 & \\
&& \name & \textbf{94.84} & \textbf{3.46} & \textbf{19.54} & \\
\bottomrule
\end{tabular}}}
\label{tab:t1}
\end{table*}

Column $IDs_{att}$ is the number of the attackable trajectories on the given tracker and dataset. From the experimental results, we can observe that:
\begin{itemize}
\item  Compared with the baselines, \name achieves the highest $Succ.$ on the three datasets with fewer frames $\#Fm.$ and smaller perturbations $L_2$.
\item  Generally, in the crowded pedestrian tracking scenarios (MOT20), \name can achieve a higher success rate than that in the normal scenarios (2DMOT15 and MOT17) while DetAt and Hijack do not have significant improvement. 
In summary, the results indicate the effectiveness of \name in attacking the pedestrian multi-object trackers.
\item  Hijack seems not adaptable to pedestrian multi-object trackers, especially on the crowded pedestrian datasets (MOT20). \name, by contrast, has outstanding performance on all the three benchmarks.
\item  For ByteTrack, which has better tracking ability, \name seems more stable while the success rates of other baselines decrease significantly.
\end{itemize}

In addition, we also compare the attack results under the constraints of attacked frames and $L_2$ distance (see \cref{fig:4}).
Compared with the baselines, \name yields higher success rates at certain attacked frames and $L_2$ distance constraints on the three datasets.

\begin{figure*}[!ht]
  \centering
  \begin{subfigure}[b]{0.48\linewidth}
    \includegraphics[width=1\linewidth]{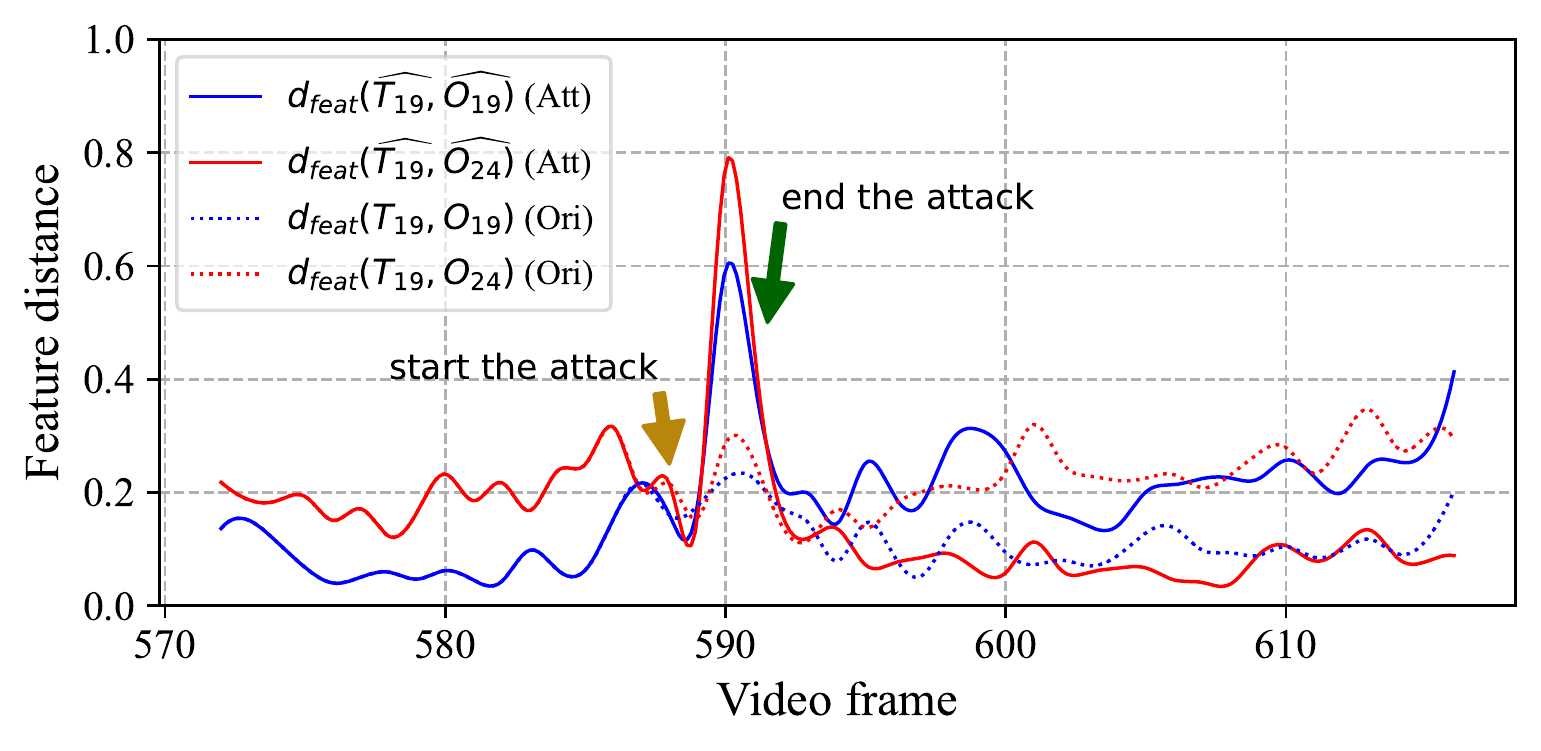}
    \caption{\footnotesize Feature distances in the video}
    \label{fig:7a}
  \end{subfigure}
  \hfill
  \begin{subfigure}[b]{0.48\linewidth}
    \includegraphics[width=1\linewidth]{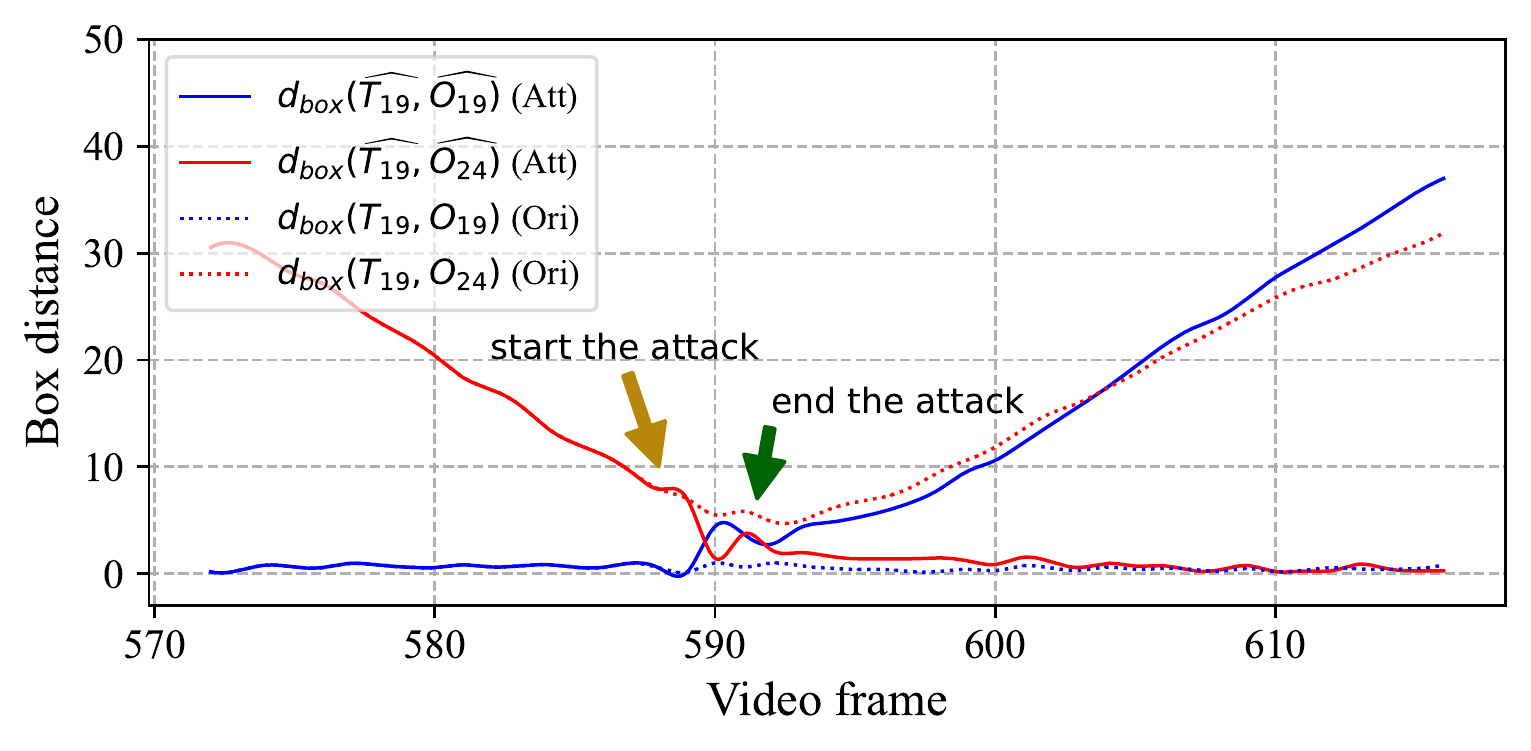}
    \caption{\footnotesize Box distances in the video}
    \label{fig:7b}
  \end{subfigure}
  \caption{\footnotesize Distances of features and boxes in the original video and adversarial video. $d_{feat}$ and $d_{box}$ represent the cosine similarity of features and the Mahalanobis distance of boxes, as presented in \cref{fm}. $\widehat{T_{id}}$ ($\widehat{O_{id}}$) and $T_{id}$ ($O_{id}$) denote the adversarial and original tracklets (objects).}
  \label{trend}
\end{figure*}

To better understand \name, we provide analysis on the attacking instance 
illustrated in~\cref{fig:f1} for the $19$-th tracklet ($24$-th tracklet is the screener object). Specifically, we plot the feature distance and box distance trends in \cref{trend}. The blue lines indicate the distance between tracklet $19$ and object with the original ID of $19$. The red lines indicate the distance between tracklet $19$ and object with the original ID of $24$. The solid lines and the dotted lines indicate the trend curve in the adversarial video and the original video. We can observe that the solid lines coincide with the dotted lines before the attack, yet after the attack, the red line and the blue line of the adversarial video switch, and tracklet $19$ is instead similar to object $24$. The original intention of updating tracklets at each time step is to adapt to variations of the tracked objects, but the update mechanism of tracklets allows us to tamper with the tracklets so that the tracker does not realize of tracking a completely different object. 


\begin{table*}[t]
\centering
\caption{Attack success rates at certain attack frames. $PP$, $CL$ and $FN$ represent the \textit{PullPush}, \textit{CenterLeaping} and \textit{Adding Failure Noise}, respectively. $n_{\#Fm.}$ presents attacking at most $n$ frames.}
\setlength{\tabcolsep}{2mm}{
\resizebox{0.8\linewidth}{!}{
\begin{tabular}[t]{c|l|ccccc}
\toprule
\multirow{2} * {Dataset} & \multicolumn{1}{c|}{\multirow{2} * {Attackers}} & \multicolumn{5}{c}{Attack success rate $\uparrow$ (\%)}\\
\cline{3-7}
& & $1_{\#Fm.}$ & $5_{\#Fm.}$ & $10_{\#Fm.}$ & $15_{\#Fm.}$ & $20_{\#Fm.}$ \\
\midrule
\multirow{4} * {2DMOT15} 
& \name w/o $PP$ & 11.4 & 61.1 & 83.5 & 90.2 & 93.2 \\
& \name w/o $CL$ & 11.2 & 67.1 & 87.6 & 91.6 & 92.8 \\
& \name w/o $FN$ & \textbf{17.8} & 66.8 & 82.9 & 85.6 & 87.2 \\ 
& \name & 14.4 & \textbf{70.5} & \textbf{87.8} & \textbf{92.1} & \textbf{93.3} \\
\midrule
\multirow{4} * {MOT17} 
& \name w/o $PP$ & ~~9.7 & 51.1 & 75.0 & 85.9 & 90.0 \\
& \name w/o $CL$ & ~~8.8 & 38.5 & 64.7 & 74.6 & 81.2 \\
& \name w/o $FN$ & \textbf{15.5} & 62.3 & 79.3 & 85.9 & 89.1 \\ 
& \name & 14.0 & \textbf{63.4} & \textbf{81.8} & \textbf{89.5} & \textbf{91.0} \\
\midrule
\multirow{4} * {MOT20} 
& \name w/o $PP$ & 13.0 & 64.6 & 86.8 & 93.2 & 94.0 \\
& \name w/o $CL$ & 11.2 & 50.8 & 74.5 & 82.2 & 84.8 \\
& \name w/o $FN$ & \textbf{20.9} & 72.3 & 88.8 & 92.9 & 93.3 \\ 
& \name & 19.4 & \textbf{75.1} & \textbf{91.5} & \textbf{95.8} & \textbf{96.5} \\
\bottomrule
\end{tabular}}}
\label{tab:t4}
\end{table*}

\subsection{Ablation Study}
\label{as}
Here we discuss the necessity of \textit{PushPull}, 
\textit{CenterLeaping}, and \textit{Adding  Failing Noise}. The comparisons are conducted on FairMOT to analyze the contribution of each component in \name. The results are shown in Table~\ref{tab:t4}. 

From the results, we can observe that: 
\begin{itemize}
\item  The performance of \name without $PP$/$CL$ is much worse than the overall \name when only a few frames are attacked (\ie, $1_{\#Fm.}$ and $5_{\#Fm.}$), especially on MOT17 and MOT20. It shows that \textit{PullPush} together with  \textit{CenterLeaping} can significantly reduce the attacked frames. 
\item  The performance of \name without $FN$ is greatly suppressed as the number of attacked frames increases. It shows that \textit{Adding Failing Noise} can make the later frames easier to be attacked, especially on 2DMOT15. However, with only one frame being attacked, \name without $FN$ seems better than the overall \name. The unsuccessfully attacked frame will be disturbed with the \textit{failing noise} for the overall \name but not for the \name without $FN$. And a frame without noise will not be seen as an attacked frame. Therefore, the attack success rate of \name without $FN$ seems slightly higher than the overall \name for only one attacked frame.
\item  On MOT17 and MOT20, the attack success rate of \name without $CL$ drops more than that without $PP$, indicating that \textit{CenterLeaping} can further improve the attack success rate in crowded scenarios. It seems that the bounding box matching mechanism of the association algorithm, which is widely used in MOT, is more vulnerable. 
\end{itemize}
In summary, the three components of \name significantly contribute to the great performance.

\subsection{Parameter Study on IoU Threshold}

\begin{figure}[!ht]
\centering
\includegraphics[width=0.9\linewidth]{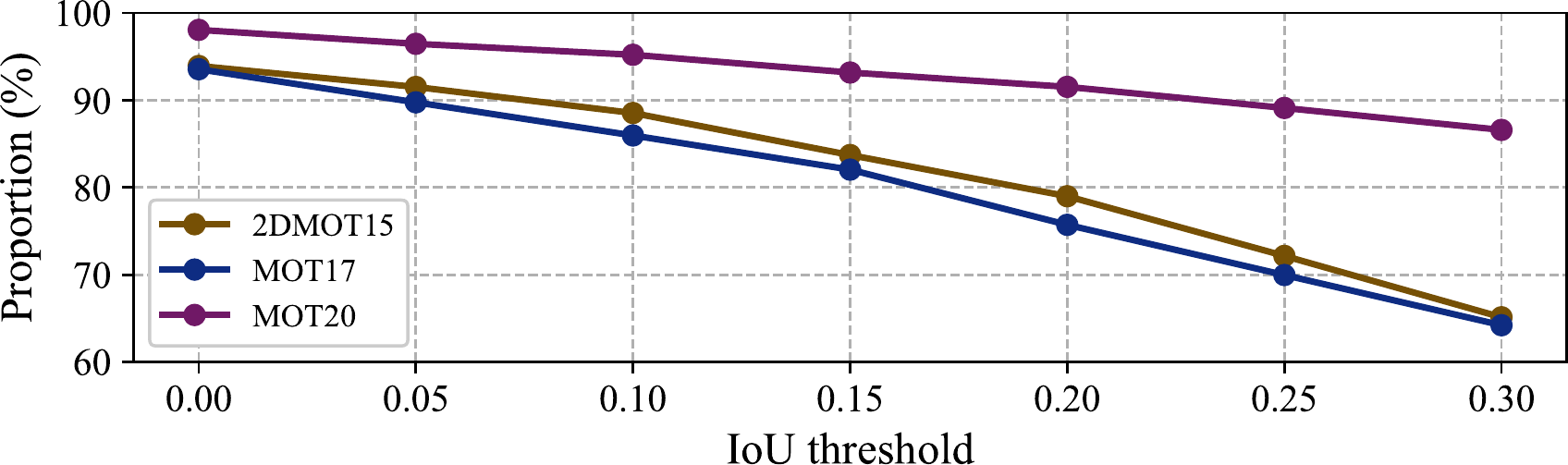}
\caption{Proportion of attackable objects.}
\label{fig:iou}
\end{figure}

One of the \name's attack conditions is that there is another object overlapping with the attack object, and the IoU is greater than the threshold $Thr_{IoU}$. It seems that the attack condition makes \name relatively limited (\ie, a few objects satisfy the attack condition). However, according to our analysis in~\cref{fig:iou}, in most scenarios, the attackable objects occupy 80\% under the experimental $Thr_{IoU}$ of 0.2; it is even as high as 90\% in the dense scenarios (MOT20). Hence, our approach can adapt to most pedestrian tracking scenarios.

\subsection{Noise Pattern}

\begin{figure}[!ht]
\centering
\includegraphics[width=0.9\linewidth]{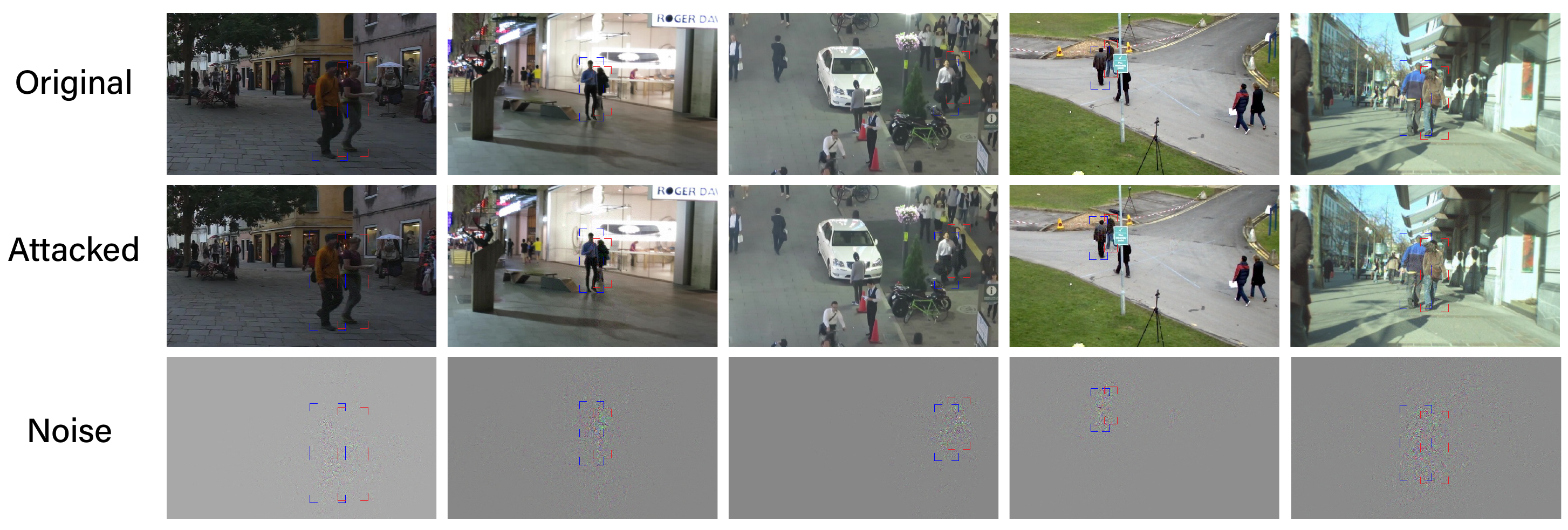}
\caption{The distribution area of noise.}
\label{fig:f8}
\end{figure}

As shown in \cref{fig:f8}, five attack tracklets are randomly selected from five videos to show the distribution area of noise.
The blue boxes and red boxes represent the attack and screener objects. The noise mainly focuses on the attack and screener objects, leaving other regions almost not perturbed. It demonstrates that \name can focus on key areas and attack 
the frames efficiently.

\section{Discussion and Future Work}
\label{sec:dis}
In this section, we firstly discuss how to extend \name to other MOT trackers. Then according to the experimental results, we provide some possible defensive directions. Finally, we propose future work in terms of robustness.

\subsection{Discussion on Attack Extension}
\label{dis1}
In this work, we study the adversarial attack against MOT trackers and choose two representative types of trackers~\cite{zhang2021fairmot,zhang2021bytetrack} to analyze and attack. As for other similarly structured MOT trackers~\cite{wang2020towards,bewley2016simple,wojke2017simple,du2022strongsort,cao2022observation,9756236} (\ie, detection-based trackers or motion-and-appearance-based trackers), \name can transplant on them with almost nothing to change. But for some different-structured models~\cite{centertrack,DBLP:journals/corr/abs-2104-00194,peng2020ctracker,Shuai_2021_CVPR}, \name should be changed to adapt to these models based on the optimization goal described in \cref{sec:pd}, which aims to switch the appearance and distance similarity of the intersecting trajectories. 

Take CenterTrack~\cite{centertrack} for example. CenterTrack takes two adjacent frames and the heat map from the previous frame as input and outputs an extra displacement prediction map compared to CenterNet~\cite{duan2019centernet}. It is similar to the optical flow method that the displacement prediction map aims to predict the displacement of the center points of the detection frame in the front and back frames:
\begin{equation}
\label{eq:displacement}
\begin{aligned} 
\mathcal{L}_{Dis}=\frac{1}{N}\sum_{i=1}^{N}\left|\hat{D}_{p_i^t}-(p_i^{t-1}-p_i^t)\right|,
\end{aligned}
\end{equation}
where $\hat{D}_{p_i^t}$ denotes the value of the 2D displacement prediction map $\hat D\in \mathbb{R}^{\frac{W}{4}\times \frac{H}{4}\times 2}$ at location of the $i$-th detected object in the $t$-th frame $p_i^t$, $p_i^{t-1}$ and $p_i^t$ are the tracked ground-truth objects. In order to perturb such models and switch the trajectories, \name needs to make some adjustments to 
adapt to CenterTrack:
\begin{equation}
\label{eq:displacement_l1}
\begin{aligned} 
\min_{\hat{D}} \left|\hat{D}_{p_i^t}-(p_j^{t-1}-p_i^t)\right| + \left|\hat{D}_{p_j^t}-(p_i^{t-1}-p_j^t)\right|
\end{aligned}
\end{equation}
and 
\begin{equation}
\label{eq:displacement_l2}
\begin{aligned} 
\max_{\hat{D}} \left|\hat{D}_{p_i^t}-(p_i^{t-1}-p_i^t)\right| + \left|\hat{D}_{p_j^t}-(p_j^{t-1}-p_j^t)\right|,
\end{aligned}
\end{equation}
where $i$, $j$ represent the attack and screener objects.

Theoretically, \name is not limited to the pedestrian Multi-Object Tracking systems. The scene of intersecting objects also appears in other Multi-Object Tracking tasks, such as autonomous driving~\cite{8443742,7368032,zhao2018multi,ess2010object}, multi-target multi-camera tracking~\cite{ristani2016performance,ristani2018features,zhang2017multi,specker2021occlusion}, \etc. Therefore, our attack method \name can also be migrated to these tasks. Although these scenes may not have as many intersecting objects as pedestrian Multi-Object Tracking has, 
the robustness of models in the case of object intersection cannot be ignored. We will continue to explore our methods on other scenarios like vehicle tracking in our future work.  

Considering that those MOT systems are not as popular and high-precision as FairMOT and ByteTrack, we did not apply \name to them yet. In this work, we argue that "when to attack" 
is more important than "how to attack", and propose an attack method called \name. The existing white-box attack methods~\cite{DBLP:journals/corr/CarliniW16a,goodfellow2014explaining,brown2017adversarial} and even black-box attack methods~\cite{papernot2017practical,yang2020learning,guo2019simple} on classification can be implemented in our method.

The ID switch is important in MOT tasks, and our method gains great achievement on online trackers. However, offline trackers can recover incorrectly matched tracklets, and how to attack the offline trackers effectively would be another possible line of future work. 

\subsection{Discussion on Possible Defense}
\label{dis2}
\name can effectively make two intersecting trajectories switch with imperceptible perturbations. How can we defend such attack? There are usually two ways to defend adversarial attacks: 1) identify the adversarial examples and deal with them (denoise or discard)~\cite{liao2018defense,pang2018max,nayebi2017biologically}; 2) improve the robustness of models~\cite{goodfellow2014explaining,sankaranarayanan2018regularizing,madry2018towards}. The first usually reduces the model accuracy and takes extra time. Based on our experimental results, we mainly discuss how to improve the robustness of trackers.

Through the above experiments, something interesting enlightens us in the defensive direction. In real-world scenarios, obviously, trackers need to distinguish different pedestrians with appearance embeddings, especially when pedestrians intersect. 
However, the process of attack in \cref{fig:7a} and the ablation study in \cref{as} show that the re-ID features of FairMOT are not discriminative enough. Take the instance in \cref{fig:f1} for example. The cosine similarity of the deep appearance features even switches without attack (see the dotted line in \cref{fig:7a}). Besides, the attack success rates of \name without \textit{PullPush} reach the same level as the overall \name as compared with \name without \textit{CenterLeaping} (see \cref{as}). It shows that only attacking the detection branch can achieve a high success rate, and the re-ID branch does not work very well when the target boxes are close to each other. Therefore, improving the robustness of re-ID branch can be used as a way of defense.

The association algorithm widely used by the state-of-the-art trackers~\cite{zhang2021fairmot,zhang2021bytetrack,DBLP:journals/corr/abs-2104-00194,du2022strongsort,wang2021multiple} also has some defects. The algorithm aims to get the global optimal with the Hungarian algorithm in each matching phase. However, for hard samples at the decision boundary, such as the close pedestrians, the algorithm is prone to errors with adversarial examples.
The tracker will continue tracking the wrong trajectories without realizing the error. 
Hence, robustness of the association algorithm should be improved to defend against such imperceptible perturbations.

\subsection{Future Work}
\label{dis3}

The study of adversarial examples is of great importance to the model robustness. In future work, we will follow up this direction and continue to explore the following problems. 

MOT often needs to handle the complex data association problem. Instead of attacking the model with a simple negative optimization method, we explore the vulnerability of MOT models from different perspectives. 
In this work, we propose a new attack method called \name under ideal scenarios, and more researches in real-world scenarios worth further exploration. 

In addition to the pedestrian multi-object tracking system, we will try to apply \name to more Multi-Object Tracking scenarios as discussed in \cref{dis1}. We believe that the robustness of all Multi-Object Tracking tasks, which exist intersecting objects, needs to be investigated and improved.

We hope that the proposed attack method could inspire more works in designing robust MOT trackers. We will also continue to design adversarial defense methods to improve the robustness of existing MOT models. As discussed in \cref{dis2}, we will also study the robustness from the defensive point of view and design an MOT system that includes redesigned feature extraction modules and post-processing algorithms.

\section{Conclusion}
\label{sec:con}

To our knowledge, this is the first work to study the adversarial attack against pedestrian MOT trackers. The proposed adversarial attack method,  \name, which consists of the \textit{PushPull} and \textit{CenterLeaping} techniques, can efficiently deceive the advanced MOT trackers at a high success rate. Exploiting the update mechanism of tracklets to attack the MOT trackers, \name also demonstrates the weakness of the re-ID branch and the association algorithm in MOT systems. Empirical experiments on standard benchmarks show that our method outperforms the existing attack methods on object detection. Considering that this is the first adversarial study on pedestrian MOT, we also provide the comparison with \textit{tracker hijacker}, which is focused on vehicle MOT, for reference. Our work shows the effectiveness and great potential of the proposed method for MOT trackers.

\section*{Acknowledgment}
This work is supported by National Natural Science Foundation (U22B2017, 62076105).

\ifCLASSOPTIONcaptionsoff
  \newpage
\fi

\bibliographystyle{IEEEtran}
\bibliography{IEEEabrv, egbib}

\end{document}